\RequirePackage{fix-cm}
\documentclass[twocolumn]{svjour3}          
\smartqed  
\usepackage{graphicx}
\usepackage[font=footnotesize]{subfig}
\usepackage{bm} 
\usepackage[usenames, dvipsnames]{color}
\usepackage{amsmath}

\usepackage{multirow}
\usepackage{cancel}
\usepackage[final]{changes}
\definechangesauthor[name={}, color=blue]{pedro}
\definechangesauthor[name={}, color=blue]{nino}
\definechangesauthor[name={}, color=blue]{bruno}
\definechangesauthor[name={}, color=blue]{jaeseok}

\definechangesauthor[name={authors}, color=blue]{authors}

\newcommand{\traj}{\mathbf{T}}
\newcommand{\homo}{\mathbf{H}}
\newcommand{\punto}{\mathbf{P}}
\usepackage{bbding}

\begin{document}

\title{Cleaning tasks knowledge transfer between heterogeneous robots: a deep learning approach}


\author{Jaeseok Kim$^{1}$ \and Nino Cauli$^{2}$ \and Pedro Vicente$^{2}$ \and  Bruno Damas$^{2,3}$ \and Alexandre Bernardino$^{2}$ \and Jos\'{e} Santos-Victor$^{2}$  \and Filippo Cavallo$^{1}$ \thanks{Parts of this manuscript were previously presented at the IEEE International Conference on Autonomous Robot Systems and Competitions (ICARSC 2018), Torres Vedras}}



\institute{J.Kim \and F.Cavallo \at
              $^{1}$BioRobotics Institute, Scuola Superiore Sant'Anna, Pisa, Italy. \\
              \email{j.kim@sssup.it, filippo.cavallo@santannapisa.it}           
           \and
           N.Cauli \and P.Vicente \and  A.Bernardino \and J.Santos-Victor \at
              $^{2}$Institute for Systems and Robotics, Instituto Superior Tecnico, Universidade de Lisboa, Portugal. \\
              \email{\{ncauli,pvicente,alex,jasv\}@isr.tecnico.ulisboa.pt} 
           \and
           B.Damas \at
              $^{2}$Institute for Systems and Robotics, Instituto Superior Tecnico, Universidade de Lisboa, Portugal. \\
              $^{3}$CINAV --- Centro de Investiga\c{c}\~{a}o Naval, Almada, Portugal. \\
              \email{bdamas@isr.tecnico.ulisboa.pt} 
}

\date{Received: date / Accepted: date}

\maketitle

\begin{abstract}

\begin{figure}
\centering
  \includegraphics[width=0.30\textwidth]{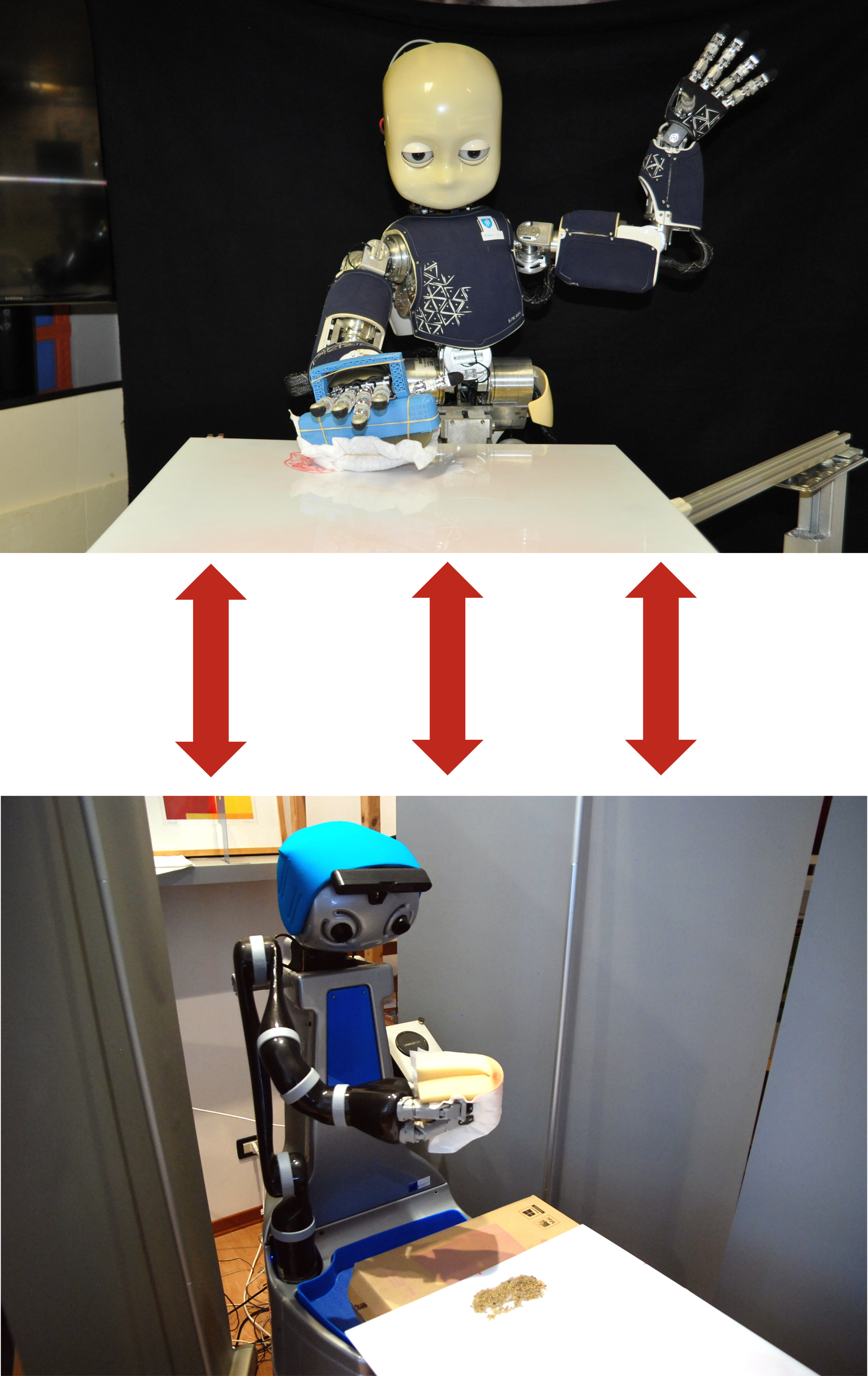}
  \caption{Picture of the Lisbon iCub robot (top) and the Peccioli DoRo robot (bottom) in their experimental setup. Our system was trained on the iCub and then tested on the DoRo.}
  \label{fig:doro}       
\end{figure}

Nowadays, autonomous service robots are becoming an important topic in robotic research. Differently from typical industrial scenarios, with highly controlled environments, service robots must show an additional robustness to task perturbations and changes in the characteristics of their sensory feedback. In this paper, a robot is taught to perform two different cleaning tasks over a table, using a learning from demonstration paradigm. However, differently from other approaches, a convolutional neural network is used to generalize the demonstrations to different, not yet seen dirt or stain patterns on the same table using only visual feedback, and to perform cleaning movements accordingly. Robustness to robot posture and illumination changes is achieved using data augmentation techniques and camera images transformation. This robustness allows the transfer of knowledge regarding execution of cleaning tasks between heterogeneous robots operating in different environmental settings. To demonstrate the viability of the proposed approach, a network trained in Lisbon to perform cleaning tasks, using the iCub robot, is successfully employed by the DoRo robot in Peccioli, Italy. 

\end{abstract}

\section{Introduction}
\label{intro}

Times where robots were relegated to controlled factory environments, with absolutely no interaction with humans, are becoming part of our past. Nowadays, robots share their working environment with us, needing the ability to handle unexpected situations, to interact with humans, and to not interfering with co-workers actions (both humans and robotic). A perfect example of robotic platforms facing these problems are the service robots. In recent years, elderly population increased exponentially around the globe, forcing the research community to find solutions to smoothly integrate senior citizens in the modern society. During assistance, caregivers are overloaded with tasks, most of them physical and repetitive. For this reason, caregivers spend a substantial amount of their time in house chores and physical assistance, overlooking social interaction with the assisted elders. Service robots able to perform house chores would relieve caregivers from a significant burden, giving them more time to spend with the elders.
Cakmak \emph{et al.} \cite{cakmak2013towards} observed that cleaning tasks are 49.8$\%$ of all chores that humans perform at home. Many mobile cleaning robots were already successfully presented in the market, but they are able to perform only highly specialized and simplified tasks, like cleaning the floor inside an apartment with a predefined behaviour. Unfortunately, typical cleaning actions, such as wiping, washing, sweeping and scrubbing, require a robot with fine manipulation abilities to be performed \cite{leidner2015classifying}. As a result, several researchers focus their attention on service robots equipped with manipulators to perform cleaning tasks \cite{okada2006vision,yamazaki2010system,liang2017dual,hess2012null,nebel2013much,dornhege2013integrated,ortenzi2014experimental,urbanek2004learning,leidner2016robotic,leidner2016knowledge,leidner2016inferring,hess2011learning,martinez2015safe,cruz2016training,devin2017deep,liu2017imitation,paxton2015incremental,kormushev2011upper,gams2016line,gams2016adaptation,pervez2018learning,silverio2015learning,alizadeh2014learning,hoyos2016incremental,rahmatizadeh2017vision,kim2018icub,pervez2017learning}.

The most direct approach to design a service robot able to perform basic cleaning actions is using classical control. Okada \emph{et al.} \cite{okada2006vision} generate a sequence of body posture to perform a sweeping motion using whole body inverse kinematics. To increase stability and avoid self collisions, Yamazaki \emph{et al.} \cite{yamazaki2010system} use the SR-inverse method to control robot's upper body during a cleaning task. Liang \emph{et al.} \cite{liang2017dual} generate a sweeping motion with both arms using full dual position control based on task-space kinematics. Hess \emph{et al.} proposed a novel coverage path planning for robotic manipulators that can clean arbitrary 3D surfaces \cite{hess2012null}. The authors suggest a generalization of the traveling salesman problem (GTSP), which transforms the surface into a graph defining a set of clusters over nodes and minimizing some cost measures. Dornhege \emph{et al.} discussed how to combine classical symbolic planning with geometric reasoning in their TFD/M (Temporal Fast Downward/Modules) planner for wiping tasks using the PR2 robot \cite{nebel2013much,dornhege2013integrated}. Ortenz \emph{et al.} suggested projected operational space dynamics that minimize joint torque and increase stability while the robot is in contact with a whiteboard during a wiping movement \cite{ortenzi2014experimental}.  Urbanek \emph{et al.} used Cartesian impedance control to create a compliant behavior of the robotic end-effector while wiping a table \cite{urbanek2004learning}. The Cartesian impedance control is extended with a compliant whole-body impedance control framework to interact with the environment using Rollin' Justin in Leidner \emph{et al.} \cite{leidner2016inferring}. The same group implemented an hybrid reasoning mechanism adding task parameterization to their whole-body control and integrating symbolic transitions to concrete cleaning actions performed using a sponge \cite{leidner2016robotic,leidner2016knowledge}. Classical control approaches are the perfect solution in case the cleaning environment is controlled,  well known a priori and does not change in time. However, if the robot faces unknown environments, it needs to adapt to unseen situations and to learn new skills. Classical controllers able to generalize to such unexpected situations are difficult to design.

In order to adapt to unknown environment and acquire new skills, cleaning robots should be able to learn from past and new experience. Using Reinforcement Learning (RL) robots can autonomously learn an approximation of optimal action policies for cleaning 
through self exploration of their action space. Hess \emph{et al.} define an efficient state transition model for wiping table using a Markov Decision Process (MDP) \cite{hess2011learning}. 
The transition function is modeled by observing the outcomes of robot's actions and then used to generate paths for cleaning table surfaces. MDP is also used to clear objects from a table in fully-observable problems with uncertainty \cite{martinez2015safe}. The same authors employ REX-D algorithm that integrates active teacher demonstration for increasing learning speed in order to sweep lentils from a plane \cite{martinez2017relational}. Interactive RL approach with contextual affordances is developed by Cruz \emph{et al.} to clean a table using state-action-reward-state-action (SARSA) \cite{cruz2016training}. In some cases the cleaning robot needs to handle high dimensional sensory data, like raw pixels data from camera images. In such situations deep reinforcement learning (Deep RL) models can simultaneously learn a desired behaviour from self exploration and extract the relevant features from raw data. 
Devin \emph{et al.} \cite{devin2017deep} developed a Deep RL object-level attentional mechanism used to control a robot in different tasks like pouring almonds in a cup or sweeping citrus from a table. Moreover, Liu \emph{et al.} proposed an imitation-from-observation algorithm used to perform various pushing and swiping actions. The model was trained both in simulation and on a real robot showing video recordings of the action from different viewpoints \cite{liu2017imitation}. RL is a powerful tool that permits to find original solution to various control problems. Anyhow its flexibility comes with some drawbacks: long training time; exploration of dangerous states and configurations (\textit{e.g.},hitting a wall during navigation or colliding with the environment during manipulation). Training in simulation can relax these problems, but performing a domain translation from simulation to the real world can be really complex.

To speed up the learning process and avoid dangerous situations, humans tend to exploit past experiences from other people (which performed  similar actions) and to imitate their movements. 
In Learning from Demonstration (LfD) algorithms robot skills are derived from observations of human demonstrations and generalized to new environments. Dynamic Movement Primitives (DMPs) and Gaussian Mixture Models (GMM) are typically used to formalize and encode unit of action as a stable dynamical system with LfD.
\\Regarding the literature on DMPs, Ghalamzan \emph{et al.} proposed an approach where DMP model and Inverse Optimal Control (IOC) are incorporated with a reward function to generate the necessary path in a new situation \cite{paxton2015incremental}. Kormushe \emph{et al.} used DMPs and upper body kinesthetic demonstrations to teach to a robot how to clean a whiteboard  \cite{kormushev2011upper}. In addition, a periodic DMP is applied to online coaching of robots in a human-robot interaction system \cite{gams2016line}. Christopher \emph{et al.} \cite{christopher1997locally} show how weights of a periodic DMP can be learned using incremental locally weighted regression (ILWR). The periodic DMP is also used with force feedback for wiping differently tilted surfaces \cite{gams2016adaptation,gams2016line}. Moreover, in Pervez \emph{et al.} \cite{pervez2018learning} task parameterized DMP (TP-DMP) is used for adaptive motion encoding to perform a sweeping task based on few demonstrations. 
\\On the GMM side, Calinon  \emph{et al.} \cite{calinon2013improving} proposed the Task-Parameterized Gaussian mixture model (TP-GMM), a technique to generalize trajectories from demonstrated ones using task parameters (frames). Silv\'{e}rio \emph{et al.} \cite{silverio2015learning} combined TP-GMM and quaternion-based dynamical systems to learn full end-effector poses of a bimanual robotic manipulator to perform a sweeping task. A similar approach using partially observable task parameters without a dynamical system is proposed by Alizadeh \emph{et al.} \cite{alizadeh2014learning}. In their work, Hoyos \emph{et al.} \cite{hoyos2016incremental} extend TP-GMM with incremental learning skill. While several TP-GMM systems have been successfully used to generate robotic cleaning motions, none of them is able to autonomously learn the task parameters from raw sensory data (\emph{i.e.} camera images). 
\\One powerful solution to extract information from raw pixel data and learn important features on the images are Convolutional neural networks (CNNs). Rahmatizadeh \emph{et al.} proposed a system able to learn multiple tasks using CNN and Long short-term memory (LSTM) networks \cite{rahmatizadeh2017vision}. The CNN plays the role of task selector and LSTM generates the robot joint command to send to the robot for cleaning small objects using a towel. Pervez \emph{et al.} \cite{pervez2017learning} proposed to use a CNN to learn the parameters of a TP-DMP directly from camera images, calling the system Deep-DMP (D-DMP). D-DMP was used to swipe different objects from a table. 

In a recent work of our \cite{kim2018icub}, we used a similar approach to learn the parameters of a TP-GMM to control a robot performing sweeping and wiping movements while cleaning a table. Two CNN based on AlexNet \cite{krizhevsky2012imagenet} are used to learn the parameters from raw input images, collecting the data through kinesthetic teaching. The main contribution compared to \cite{pervez2017learning} is the ability of the system to generate different kinds of cleaning trajectories for different kinds of dirt: sweeping cluster of lentils and wiping off marker scribbles. A common limitation of both \cite{pervez2017learning} and \cite{kim2018icub} is the need to retrain the system for different camera positions, environment to clean and robot to use. To solve these issues, in our last published work \cite{cauli2018autonomous} we project the robot camera images into a canonical bird-view camera plane and we augment the dataset changing the illumination, shifting the images and applying Perlin noise \cite{perlin1985image} to the background. One CNN is used to directly predict means and covariances of a GMM, using GMR to obtain an estimation of the desired cleaning trajectory. After being trained with right arm's kinesthetic demonstrations, the robot was able to transfer his knowledge sweeping and wiping different kind of dirt using the left arm.

In this paper, we extend the works presented in \cite{kim2018icub} and \cite{cauli2018autonomous} using a CNN/TP-GMM system, trained on a dataset collected on the iCub robot in Lisbon-Portugal, to control the Domestic Robot (DoRo) in Peccioli-Italy while cleaning a table. 
\added[]{This type of transfer learning of a given task across different domains is known as transductive transfer learning \cite{pan2010survey} and also referred to as multi-robot transfer learning in the robotics community~\cite{helwa2017multi}. }

\subsection{Contributions}
\label{sub:contrib}

\added[]{To successfully transfer the knowledge gathered while training the iCub robot to the controlling task on the DoRo robot it is essential to collect a set of features that are invariant across domains, which is done using the techniques described in this paper, \emph{e.g.}, viewpoint invariance using a virtual camera approach and dataset augmentation using Perlin noise, image translation and change in illumination.
In addition, we perform a systematic analysis of the number of kinesthetic demonstration needed to successfully swipe and wipe off a table from lentils and marker scribbles, studying which type of data augmentation is more appropriate for our task. Thanks to this analysis we significantly reduced the high amount of kinesthetic demonstrations used in both \cite{kim2018icub} and \cite{cauli2018autonomous}.}

\added[]{
In this paper, we adopted the same transformation to a canonical virtual camera used in \cite{cauli2018autonomous}. We also used the kinesthetic demonstrations collected in \cite{cauli2018autonomous} to create the new augmented dataset. Differently from \cite{kim2018icub} and \cite{cauli2018autonomous}, we calculated the hand orientations analytically. The orientations extracted from the kinesthetic demonstrations performed on the iCub are optimal for that particular robot and do not generalize well on the DoRo.}

\added[]{
The main focus of this paper is to transfer the knowledge acquired by the iCub to the DoRo robot. For this reason, both the networks presented in \cite{kim2018icub} and \cite{cauli2018autonomous} could have been used. The solution proposed in the former is more structured compared with the end-to-end solution of the later. Indeed, the combination of a CNN and a TP-GMM produces robust results and makes the network of \cite{kim2018icub} more suitable for this work.}

\added[]{
The main contributions of this paper are:}
\begin{enumerate}
  \item \added[]{\textbf{Demonstration of successful transferring of knowledge from a robot to another:} CNN and TP-GMM trained on the iCub are used to control the DoRo robot. To achieve this, three key generalization mechanisms are used:}
  \begin{enumerate}
    \item \added[]{Geometric image transformation (bird-eye view) to cope with different robot camera geometry (pose and calibration parameters). This was already proposed in \cite{cauli2018autonomous}.}
    \item \added[]{Data augmentation to cope with different robot camera photometric properties (brightness, contrast, noise, color balance) and background clutter. This was already proposed in \cite{cauli2018autonomous} but extended in this paper with a dual Perlin noise strategy.}
    \item \added[]{End-efector orientation computed analytically to better adapt to different robot kinematics. This is a new method proposed in the current paper.}
  \end{enumerate}
  \item \added[]{\textbf{Finding an optimal number of demonstrations needed to learn a cleaning motion:} CNN are trained with different number of kinesthetic demonstrations in order to detect a good compromise between size of the dataset and performance of the network.}
  \item \added[]{\textbf{Proving the importance of domain randomization in our scenario:} Augmenting the dataset adding random Perlin noise to the background of the images is fundamental to generalize from iCub to DoRo.} 
\end{enumerate}

\subsection{Outline}
\label{sub:outline}

The paper is organized as follows. Section II summarizes
the proposed approach, describing the canonical virtual camera projection, showing the CNN architecture and giving a brief introduction to TP-GMM. Section III shows the experimental setup, while in Section IV the experimental results achieved are presented. Section V concludes the paper and
gives some directions for further research.

\section{Proposed approach}

\begin{figure*}[t]
\centering
\includegraphics[width=0.95\textwidth]{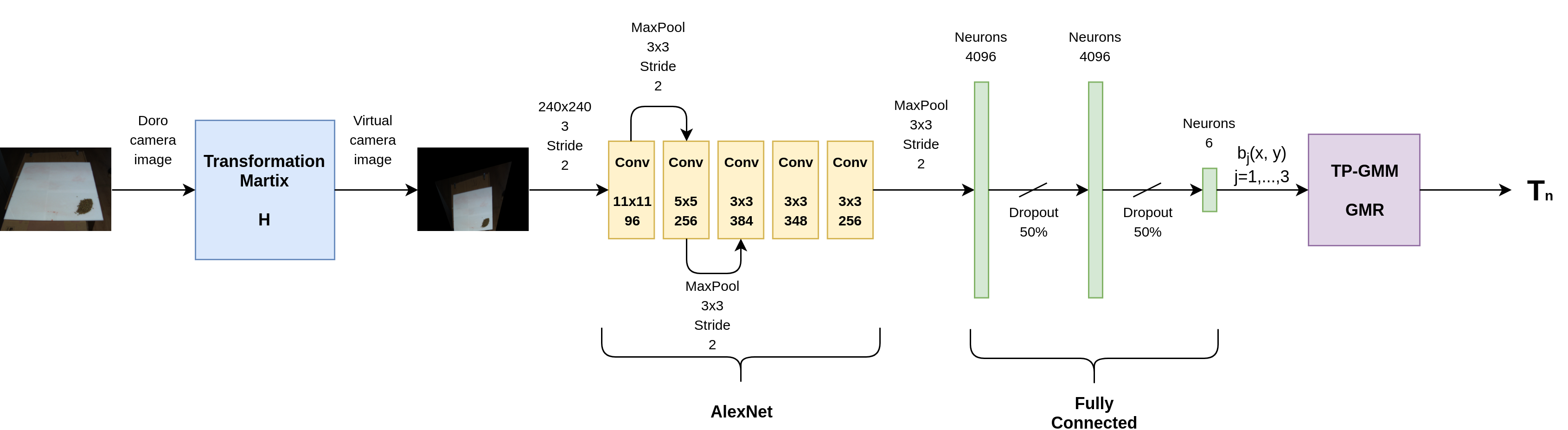}
\caption{System Architecture. Images from the robot's camera are transformed to virtual bird-view images. The virtual images are passed to a CNN that predicts initial, intermediate and final positions ($b_1$, $b_2$ and $b_3$). From the resulting TP-GMM the expected hand's trajectory is computed using GMR algorithm.}
\label{fig:sysArch}
\end{figure*}

The goal of this paper is to transfer the knowledge acquired by the iCub robot in Lisbon, during a kinesthetic demonstration of a cleaning task, to the DoRo robot in Peccioli. Two different cleaning movements are taught to the iCub in order to clean a table: a sweeping motion to remove lentils from the table and a wiping motion to clean marker scribbles. The robot holds a sponge in its hand to perform the cleaning trajectories. In order to generalize to different robot camera positions and table heights, camera images are transformed to a canonical virtual image plane, similarly to what has been done in \cite{cauli2018autonomous}. The canonical virtual camera is placed at a fix distance from the table, right on top of it, generating a bird-view image. Specific sizes and positions of objects placed on the table correspond to particular sizes and positions in the virtual image plane. From the virtual images, the robot estimates the correct cleaning hand trajectories using the same architecture introduced in our ICARSC 2018 paper \cite{kim2018icub}: a CNN estimate the initial, intermediate and final positions of the desired trajectory ($\traj(n)$, $n = 0, \dots, 200$) used to create the parameters of the TP-GMM (reference frames $\bm{X}_j = \{\bm{A}_{j},\bm{b}_{j}\}$, $1\leq j \leq 3$); GMR algorithm is used to estimate the desired trajectory $\traj(n)$ from a TP-GMM defined by the reference frames $\bm{X}_j$. The only difference from \cite{kim2018icub} is the absence of the CNN to calculate the orientations of the initial, intermediate and final reference frames. High variation in the orientations of human demonstrations make impossible for the CNN to precisely predict the initial, intermediate and final orientations. To overcome this problem, in this paper we decided to analytically calculate the reference frames orientations from the reference frame positions predicted by a single CNN. Figure \ref{fig:sysArch} shows the complete system architecture.

\color{blue}

\color{black}

\subsection{Virtual camera} \label{sec:virtualcam}
The naive approach to use directly the unprocessed images taken from the robot cameras as input of the CNN has one important drawback. In real scenarios a robot approaches the table to clean it from different positions and with different head configurations. The pose of the camera plane relative to the table plane during cleaning changes dynamically. This means that the CNN should intrinsically learn this spatial correlation directly from images. The task is not easy, and several demonstrations with different camera angles covering most of the possible configurations should be recorded. In our case this implies tens of thousands kinesthetic demonstration, something impossible to generate. One solution can be to fix a specific camera/table pose during training and place the robot always in the same configuration and position relative to the table during test. This is not a realistic scenario and, even if reasonable as proof of concept, such a system is not usable in real life. 

In this paper we decided to use \added[]{the approach adopted in \cite{cauli2018autonomous}}: apply a homographic transformation $\homo$ to the robot camera plane in order to project it to a canonical virtual camera plane facing downward and placed right above the table at a fixed position. 

This post-processing guarantees input images to be always taken from the viewpoint of the same canonical virtual camera, releasing the CNN to learn the geometric transformation between image plane and table plane. In order to generate the virtual camera image, the robot calculates an homographic transformation $\homo$ from the robot camera plane to the virtual camera plane each time a new image is received:
\begin{equation}
\label{equ:hom}
z \left [
  \begin{array}{ccc}
  x_{v} \\
  y_{v} \\
  1 
  \end{array}
\right ] 
= \homo 
\left [
  \begin{array}{ccc}
  x_r \\
  y_r \\
  1 
  \end{array}
\right ]
\end{equation}
where $\punto_{r} = (x_{r},y_{r})$ and $\punto_{v} = (x_{v},y_{v})$ are pixel coordinates in the real camera frame and virtual camera frame respectively and $z$ is an arbitrary, non-zero scale factor. 

Homography matrix $\homo$ is calculated using the projection on both robot camera plane and virtual camera plane of 4 point laying on the table $\punto(i) = (x(i), y(i), h_t)$, $i=1,\dots,4$, where $h_t$ is the table height in the robot reference frame. To obtain the 4 points $\punto(i)$ the robot places his hand in 4 distinct positions on the table and uses his kinematics to extract the point planar coordinates $(x(i), y(i))$ and the high of the table $h_t$. The projection of $\punto(i)$ on the robot camera plane $\punto_{r}(i)$ is obtained using the body-eye forward kinematics and the intrinsic parameters of the camera:
\begin{equation}
\label{equ:fr}
z
\left [
  \begin{array}{ccc}
  x_r(i) \\
  y_r(i) \\
  1 
  \end{array}
\right ] = K[I|\mathbf{0}] \cdot \tau_O^{eye}(q) \cdot \left [
  \begin{array}{cccc}
  x(i) \\
  y(i) \\
  h_t \\
  1 
  \end{array}
\right ]
\end{equation}
where $\tau_O^{eye}(q)$ denotes the Denavit-Hartenberg matrix from the robot reference frame $O$ to the camera reference frame, $I$ is a 3x3 identity matrix and $\mathbf{0}$ is a 3x1 vector of $0$s.

To calculate the projections on the virtual camera plane $\punto_{v}(i)$ we use the following function relating the robot reference frame $O$ and the virtual camera image frame:
\begin{equation}
\label{equ:fv}
\punto_{v}(i) = \left (
  (y(i) + 2/3)h,\quad(x(i) + 1)h
\right ) 
\end{equation}
where $h$ is a scaling factor from pixel to meters that correspond to the height of the virtual camera image expressed in pixels. All points in 3D space are expressed on the iCub reference frame $O$ placed near the hips of the robot. Ideally, the calibration process (\textit{i.e.},the robot touching 4 different positions on the table and extracting the points $\punto(i)$) must be repeated every time the robot approaches a new table.

The procedure described above was used to generate the images of our dataset. In the case of DoRo, the head stays still during the entire cleaning experiment. For this reason we decided to skip equation \ref{equ:fr} and select directly, from the camera image, the pixels corresponding to the points touched by the robot during calibration. 

\subsection{Data Augmentation} \label{sec:augmentation}
\begin{figure}
\centering
  \includegraphics[width=0.48\textwidth]{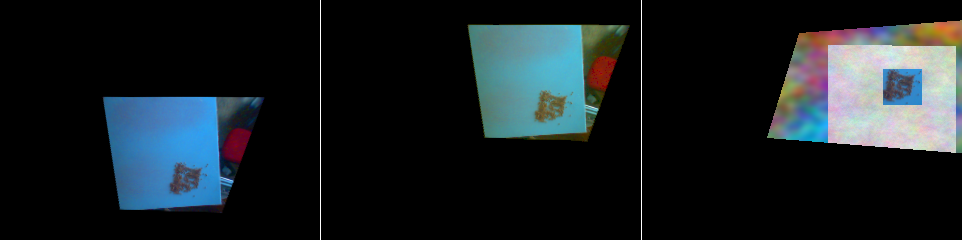}
  \caption{Examples of data augmentation. Left: original dataset image. Center: same image with change in illumination and translation. Right: same image with Perlin noise table and background.}
  \label{fig:aug}       
\end{figure}
A second step necessary to achieve the transfer of cleaning capabilities from the iCub to the DoRo, was to perform a data augmentation on the original dataset of 659 elements. Specifically, three kind of data augmentation were performed (see Fig. \ref{fig:aug} for an example):
\begin{enumerate}
  \item \textbf{Changes in illumination:} a random value in between -0.15 and 0.15 is added to each RGB channel.
  \item \textbf{Pixels and trajectories translation:} all pixels are translated in x and y with a random value ensuring that the dirt stays always visible inside the image. A correspondent shift in meters is then applied to the 200 trajectory points.
  \item \textbf{Substituting background and table with Perlin noise:} we cropped the dirt from the image and substitute the background and the table with Perlin noise. Both projected field of view and table shape were slightly randomized. 
  \added[]{This data augmentation strategy is important to perform the transfer learning between the two robots, since the background is different.}
\end{enumerate}

To create the final dataset we augmented 10 times the original one applying changes in illumination and translations, plus more 10 times applying Perlin noise.

\added[]{The original 659 elements are the same used to create the dataset in \cite{cauli2018autonomous}. The difference is in the augmentation with the Perlin noise. Instead of substituting the entire background with a Perlin noise texture as in \cite{cauli2018autonomous}, we added two different Perlin noise textures, one for the table and one for the background (see Figure \ref{fig:aug}). In this way we are able to keep a basic structure of the environment inside the randomization.}

\subsection{End-effector Control}

In our previous work \cite{kim2018icub}, we used two CNNs to obtain three reference frames $\bm{X}_j, 1\leq j \leq 3$ used as task parameters for the TP-GMM (one network for orientations $\bm{A}_{j}$ and the other one for positions $\bm{b}_{j}$ of the end-effector).
 \replaced[]{Due to the high variability in kinesthetic demonstrations the orientation estimation error was high when evaluated on a test set comprising demonstrations that also presented a large variability in the orientation of the reference frames. As a consequence, in this work we decided to simplify the architecture and to use only one CNN to estimate the initial $\bm{b}_{1}$, intermediate $\bm{b}_{2}$ and final $\bm{b}_{3}$ positions, obtaining the rotation matrices $\hat{\bm{A}_{j}}$ directly from the estimated positions $\hat{\bm{b}_{j}}$.}{Unfortunately, the performances of the CNN used to predict the orientations were worst than expected (mean error of 0.7621 radians) due to the high variability in kinesthetic demonstrations. In this paper it was decided to use only one CNN to estimate the initial $\bm{b}_{1}$, intermediate $\bm{b}_{2}$ and final $\bm{b}_{3}$ positions, and to obtain an estimation of the rotation matrices $\hat{\bm{A}_{j}}$ directly from the estimated positions $\hat{\bm{b}_{j}}$.} More precisely, $\hat{\bm{A}_{j}}$ are 2D rotation matrices of angles $\theta_{j}$ calculated as follow:
\begin{equation}
\label{equ:orig}
\begin{array}{ccc}
sin(\theta_{j}) = \frac{y_{dist, j}}{\sqrt{{x_{dist, j}}^{2} + {y_{dist, j}}^{2}}} & \\
& \qquad 1\leq j \leq 3\\
cos(\theta_{j}) = \frac{x_{dist, j}}{\sqrt{{x_{dist, j}}^{2} + {y_{dist, j}}^{2}}} & 
\end{array}
\end{equation}
where $x_{dist, j}$ and $y_{dist, j}$ are the differences in $x$ and $y$ axis between initial and intermediate points in case of $j=1$, and between intermediate and final points in case of $j=2,3$.
\added[]{Calculating the end-effector orientation this way is more flexible than extracting it from the kinesthetic data. The motion constraints of iCub and DoRo are quite different, same for the orientations achievable by each robot in a specific position. The orientations recorded moving the iCub's hand are biased by the kinematics costraints of the robot. Using equation \ref{equ:orig} to calculate the 2D rotation matrices better generalize on a different robot with different constraints.}

\subsection{Convolutional neural network}

The network architecture used in this paper is the same as in \cite{kim2018icub}. The CNNs architecture was devised based on the AlexNet~\cite{krizhevsky2012imagenet} model changing only the output layer. In the proposed networks, the 1000 nodes output layer of the AlexNet is replaced with a fully connected one with 6 nodes. The outputs of the network are the x and y Cartesian coordinates of the three reference frames. The network takes as input a 3 channel (RGB) image resized to a dimension of 240x240 pixels.
The network has a total of 8 layers: 5 convolutional layers and 3 feedforward fully connected layers.
Fig.~\ref{fig:sysArch} depicts a detailed description of the network architecture. In order to train the network we generated a dataset of virtual camera images and trajectories using 659 kinesthetic demonstrations of wiping and sweeping movements. During training we minimize the mean square error (MSE) between the network outputs and the initial, intermediate and final positions ($x$ and $y$) of the hand trajectories. See Section~\ref{sec:Dataset} for a detailed description of this dataset.

\subsection{Task parameterized Gaussian mixture model and Gaussian mixture regression}

The use of a Gaussian Mixture Model to represent a set of trajectories performed by a human demonstrator is an efficient way of representing such demonstrations in a compact way, as all data points will be represented by a mixture of Gaussians that encompasses the average demonstrated trajectory, together with a corresponding variability. Under the LfD paradigm each demonstrated trajectory $m$ from a set of $M$ demonstrations consists of a set $T_m$ vectors of dimension $D+1$, each vector $\bm{\xi}_n$ containing the observed task space variables $y_n$ and the corresponding time $t_n$, for $1 \leq n \leq N$ and $N = \sum_{m = 1}^M T_m$.
By training a GMM with $K$ components on this data set and then conditioning the resulting GMM on the time variable $t_n$ one can obtain an average trajectory as a function of time, to be performed by the robot: this is known as Gaussian Mixture Regression (GMR)~\cite{ghahraiuani1994solving}.

Task Parametrized GMM is an extension of GMM that allows the extrapolation of skills to different regions of the task space or to make such learned skill depend on a set of external variables, \emph{e.g.}, a set of via points for the trajectory of the end-effector that the robot must reach in succession. This is done in~\cite{calinon2013improving} by considering a set of auxiliary frames of reference that define initial, intermediate and final points for the trajectory to perform. Each frame of reference $j$, $1\leq j \leq P$, is represented by its origin $\bm{b}_{n,j}$ and rotation matrix $\bm{A}_{n,j}$.

We use the same framework as in~\cite{calinon2013improving} to learn to perform a cleaning movement from human demonstrations. Differently from that work we consider fixed frames of reference for each demonstration, as these are automatically calculated for each demonstration, and so we make its parameters depend solely on demonstration index $m$, \emph{i.e.}, we use origin $\bm{b}_{m,j}$ and rotation matrix $\bm{A}_{m,j}$ instead. The Expectation-Maximization (EM) algorithm~\cite{mclachlanEMalgorithm} is used to train the TP-GMM: the likelihood function to maximize is $p(\bm{\overline{\xi}}|\cdot) = \prod_{n=1}^N p(\bm{\xi}_n|\cdot)$, with
\begin{equation}
    \label{eq:gmm}
	p(\bm{\xi}_n|\cdot) = \sum_{i=1}^K \pi_i p(\bm{\xi}_n|i)\;,
\end{equation} 
where $\pi_i$ are the mixture proportions, $\bm{\overline{\xi}} = \{\bm{\xi}_n\}$ and $ p(\bm{\xi}_n|i)$, the probability of mixture component $i$ generating data point $\bm{\xi}_n$, is given by the joint distribution w.r.t. reference frames,
\begin{equation}
	p(\bm{\xi}_n|i) = \prod_{j=1}^P p(\bm{\xi}_n|j,i)\;.
\end{equation}

With $\bm{\xi}_n|j,i \sim \mathcal{N}\left(\bm{A}_{m,j}\bm{Z}_{i,j}^\mu + \bm{b}_{m,j}, \bm{A}_{m,j}\bm{Z}_{i,j}^\Sigma\bm{A}_{m,j}^T\right)$, this follows a normal distribution $\bm{\xi}_n|i \sim \mathcal{N} (\bm{\mu}_{m,i},\bm{\Sigma}_{m,i})$, with
\begin{equation}
  \label{tmgmm_2}
  \Sigma_{m,i} = \bigg(\sum_{j=1}^P (\bm{A}_{m,j}\bm{Z}_{i,j}^\Sigma\bm{A}_{m,j}^{T})^{-1}\bigg)^{-1} \quad\text{and}
\end{equation}
\begin{equation}
  \label{tmgmm_3}
  \mu_{m,i} = \Sigma_{m,i} \sum_{j=1}^P (\bm{A}_{m,j}\bm{Z}_{i,j}^\Sigma\bm{A}_{m,j}^{T})^{-1}(\bm{A}_{m,j}\bm{Z}_{i,j}^\mu + \bm{b}_{m,j})\;,
\end{equation}
where index $m$ is the demonstration corresponding to data point $\bm{\xi}_n$.

Parameters $\bm{Z}_{i,j}^\mu$ and $\bm{Z}_{i,j}^\Sigma$ correspond to the mean vectors and covariance matrices describing a GMM for the data as seen from each frame of reference; together with $\pi_i$ they correspond to the parameters to be learned using the EM algorithm. The most relevant feature of this approach is that in this process different weights are assigned to different frames of reference, according to the current time of the reproduction, thus effectively capturing the most relevant features of the human demonstrations. These correspond to some invariance of the demonstrations as seen from each frame of reference, encoded in a low variance estimate for the task space variables, taken from the corresponding GMM.

The EM training procedure finds, in the E-Step, responsibilities
\begin{equation}
	\gamma_{n,i} = \frac{\pi_i p(\bm{\xi}_n|i)}{\sum_{k=1}^K \pi_k p(\bm{\xi}_n|k)}
\end{equation}
and uses these values to update estimates for parameters $\bm{Z}_{i,j}^\mu$, $\bm{Z}_{i,j}^\Sigma$ and $\pi_i$ in the M-Step (for more details please refer to~\cite{calinon2013improving}). After learning, given a new set of frames of reference $\bm{X}_j = \{\bm{A}_j, \bm{b}_j\}$, provided by the neural network from the test image, a trajectory $\bm{T}_n$ is generated in the task space by conditioning the distribution $p(\bm{\xi}|\cdot)$ on the time variable $t_n$, using~(\ref{eq:gmm}), ~(\ref{tmgmm_2}) and ~(\ref{tmgmm_3}).

\begin{figure*}
\centering
\subfloat[Training Loss]{\includegraphics[trim=0.0cm 0cm 0.8cm 0cm, clip=true,width=0.46\linewidth]{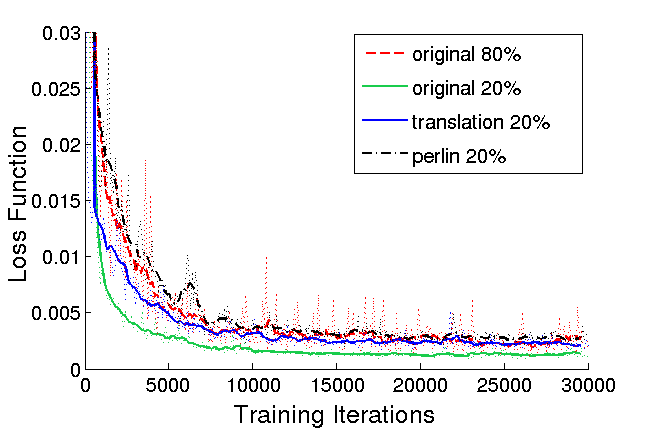}}
\subfloat[Validation Loss]{\includegraphics[trim=0.0cm 0.0cm 0.8cm 0cm, clip=true,width=0.46\linewidth]{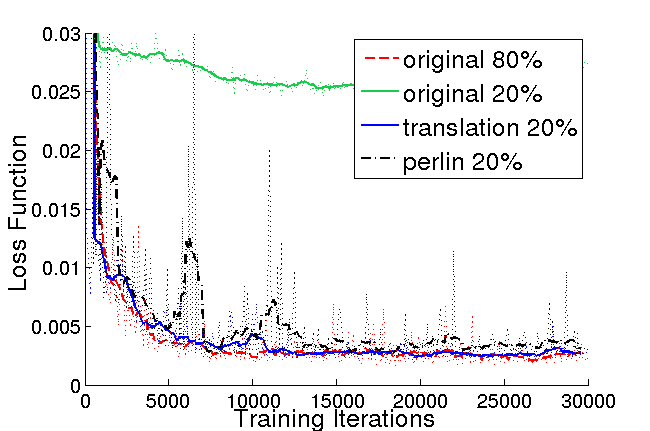}}

\caption{Training and Validation Loss of the 4 used Network. (Best viewed in color)}\label{fig:lossNetworks}
\end{figure*}

\section{Experimental Setup}

In this section, we will describe the two robots used on this work, the dataset collected and how the system was initialized. The iCub robot was used to collect the dataset using zero torque controllers to perform the kinesthetic teaching demonstrations, while the DoRo robot was used to test the system in a real world scenario accessing the transferring capabilities of the proposed architecture.

\subsection{iCub Description}

The iCub humanoid robot \cite{metta10iCub} has 53 motors that move the hands, arms, head, waist and legs. Regarding the sensory capabilities, it has a stereo vision system (cameras in the eyeballs), touch (tactile fingertips and artificial skin), vestibular sensing (IMU on top of the head) and proprioception (motor encoders and torque sensors),  which are major features that allowed us to record the dataset used in this article.

\subsection{DoRo Description}

The domestic robot (DoRo) \cite{cavallo2014development} is a service robot moved by a SCITOS G5 mobile platform (developed by Metralabs). A Kinova Jaco arm (6 DoF manipulator integrated 3-DOF hand) is mounted on the right side to perform manipulation tasks. On board are present a front laser (SICK S300) and a rear laser (Hokuyo URG-04LX) to view and avoid obstacles and perform self-localization. A pan-tilt system is installed on the head with two high-res cameras equipped with different lenses, and an Asus Xtion Pro RGB-Depth camera used for object detection. The eyes are equipped with multicolor LEDs and a speaker is used to interact with the users.

\subsection{Collecting the Dataset}
\label{sec:Dataset}
In this work we used a dataset composed by 659 kinesthetic demonstrations collected in \cite{cauli2018autonomous} changing the data augmentation strategies \added[]{(see Section \ref{sec:augmentation} for more details)}. To collect the dataset we placed the iCub robot, holding a sponge on its right hand, in front of a white table of size 50x50 cm. For each demonstration some dirt was placed on the table (lentils clusters or marker scribbles). A human guided the iCub right hand cleaning as much as possible of the dirt spot with a specific motion for each dirt type. The inputs of the dataset are images of the dirty table recorded from the iCub right eye before to perform each kinesthetic demonstration and the labels are the right hand 2D trajectories in $x$ and $y$ of the robot reference frame. Each trajectory consists of 200 elements. This dataset was then augmented using the procedure explained in Section \ref{sec:augmentation} resulting in a new dataset \replaced[]{20 times bigger}{of 13839 elements}. In order to train the CNN we extracted from the trajectories the first, intermediate and final points.

\subsection{System initialization}
\begin{figure}[t]
\centering
\includegraphics[width=0.40\textwidth]{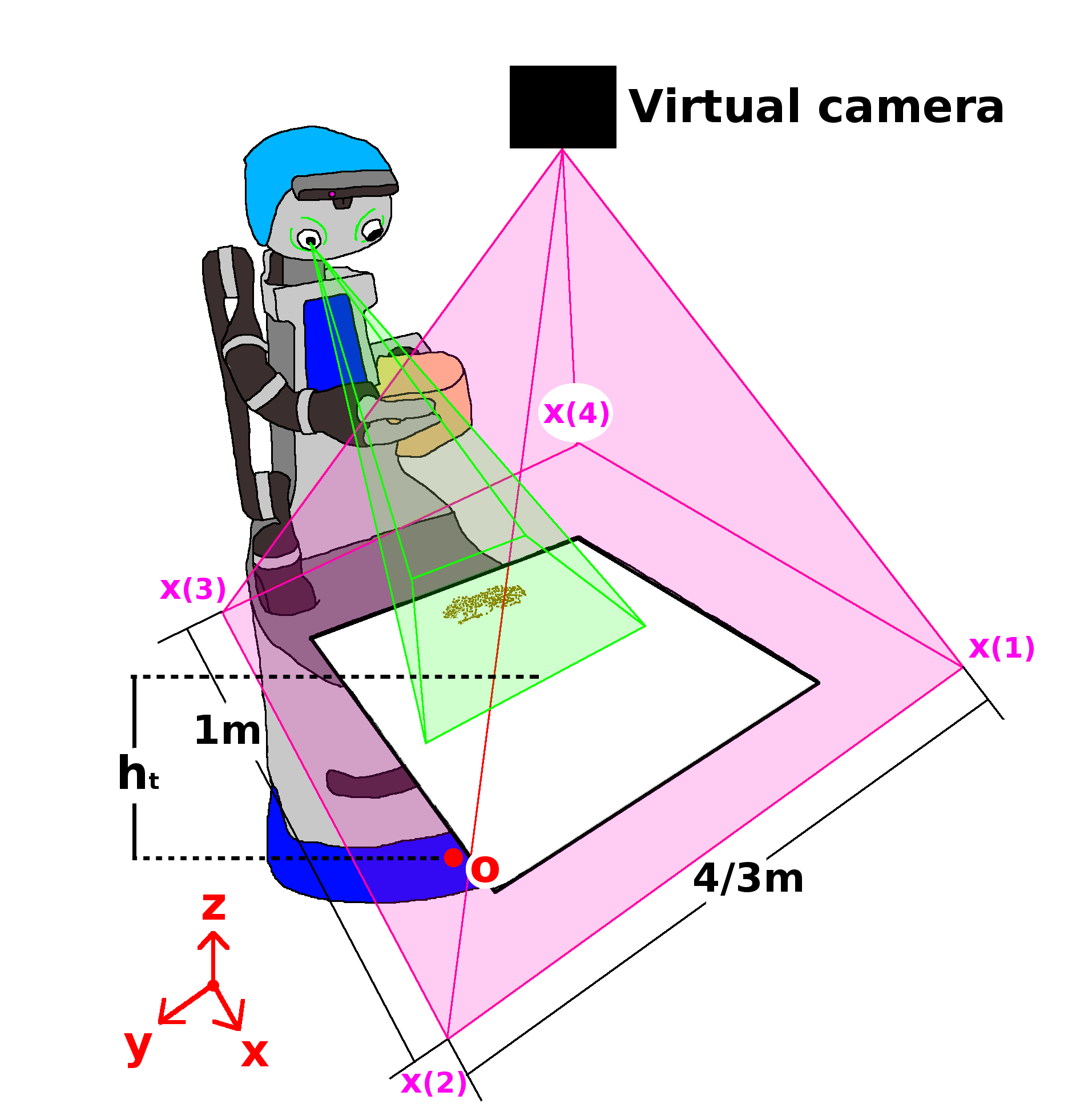}
\caption{The workspace used in our experiments. The Doro robot is placed
in front of a table. $h_ {t}$ is the height of the table expressed in the robot reference frame $O$. A bird-view virtual camera is placed on top of the table.}
\label{fig:work}
\end{figure}

\added[]{
Fig.\ref{fig:work} depicts the workspace used in the experiments with DoRo.}
The system setup was initialized with the DoRo robot with its head and arm in a pre-defined position. Indeed, this initial position can be changed without losing generality and without the need of re-training the whole system. Moreover, a 50x50~cm table was placed in front of the robot at different heights: 67~cm, 70~cm and 79~cm. The z-coordinate used when generating the final end-effector trajectory was pre-defined matching the measured table height. 
Although, the table height was measured by the experimenter in the DoRo case, it is possible to use a calibration routine exploiting, for instance, touch or torque sensors on the hand to feel the table and extract this information.
Furthermore, the virtual camera was calibrated placing the robot arm on the table in four different positions as described in the Virtual Camera section (see Section~\ref{sec:virtualcam}) using a joy-stick controller. 
\added[]{
To obtain similar images to the one collected on the iCub, we used the same table relative position to the virtual camera adopted in \cite{cauli2018autonomous}. As long as the proportion between meters and virtual camera pixels is the same as in the iCub generated dataset, the size of the table is not relevant. Due to the DoRo kinematic structure, we adopted a top grasp configuration to hold the sponge. The use of task space coordinates during the learning phase provides invariance to a particular robot kinematics model, as long as the 3 reference frames provided by the CNN are sufficiently far from the robot kinematics singularities. In our experiments, this is achieved by constraining the region of the table to be cleaned to be in the robot reachable space.}

The collected dataset was divided in two sets: i) training and ii) validation. The validation set was defined as 20\% of the original dataset (\textit{i.e.}, 20\% of 659 human demonstrations) and the training set was build selecting the remaining 80\% and performing data augmentation.
The six position labels (\textit{i.e.}, the reference frames for the TP-GMM - $\Vec{X}_j$) were normalized to improve the learning and the mean image of the training set was also calculated and subtracted from the input image on each training example. The network was implemented using Caffe \cite{jia2014caffe} and trained with the Adam optimizer with a fixed learning rate of 0.001 and dropout with ratio 0.5 on the first two fully-connected layers (see Fig.~\ref{fig:sysArch}).
Moreover, the training process used a batch size of 80 and was stopped after 30000 iterations (about 1000 epochs).

\subsection{Evaluation method}
\label{sub:errorMetrics}

The system was tested placing a dirt spot (marker or lentils) on the table and making the robot clean it in 5 repetitions without human intervention. The evaluation of the cleaning task is defined according to the different type of dirt presented on the environment. For the case of marker scribbles, the percentage of dirty area $m_1(r)$ after each repetition was calculated:
\begin{equation}
\label{equ:metric1}
m_1(r) = \frac{A(r)}{A(1)}100 , \qquad r = 1, \dots, N_{r},
\end{equation}
where $A(r)$ is defined as the dirty area in pixel at repetition $r$, and $N_r = 5$ is the number of repetitions.


For the lentils case, the performance is evaluated using the metric $m_2(r)$, which is defined by the following expression:
\begin{equation}
\label{equ:metric2}
m_2(r) = \frac{D(r)}{D(1)}100 , \qquad r = 1, \dots, N_{r},
\end{equation}
where $D(r)$ measures how far the dirt is from the target position.
$D(r)$ is defined as:
\begin{equation}
D(r) =  \sum_{i=1}^{N_p} I(i) \times \sqrt{(i - o)^T(i - o)},
\end{equation}
where $I(i)$ is an indicator function which identifies the dirty pixels, $o$ is the  bottom-right corner of the table (the target position) expressed in pixels and $N_p$ is the total number of pixels in the input image. 

To calculate the dirty area in the images, we used a post-processing phase where a color (RGB-based) segmentation was performed.

\newcommand{\rs}{\scriptsize}
\begin{table*}
\centering
    \resizebox{\textwidth}{!}{
        \begin{tabular}{l c c c c c c c c c c c c c c c c c}
                        & \multicolumn{3}{c}{\textbf{80\%}} & & \multicolumn{3}{c}{\textbf{50\%}} & & \multicolumn{3}{c}{\textbf{20\%}} & & \multicolumn{3}{c}{\textbf{10\%}} \\
                        & \multicolumn{3}{c}{\rs(527 original samples)} & & \multicolumn{3}{c}{\rs(330 original samples)} & & \multicolumn{3}{c}{\rs(132 original samples)} & & \multicolumn{3}{c}{\rs(66 original samples)} \\
        
        \cline{2-4}
        \cline{6-8}
        \cline{10-12}
        \cline{14-16}
        
                        & \textbf{O}     & \textbf{T}     & \textbf{P }    & &\textbf{ O}   & \textbf{T}     & \textbf{P}     & & \textbf{O}   & \textbf{T}     & \textbf{P}     & & \textbf{O}  & \textbf{T} & \textbf{P} \\ 
       \rs Effective Training Samples   &\rs527  &\rs5797 &\rs11067& &\rs330&\rs3630 &\rs6930 & &\rs132&\rs1452 &\rs2772 & &\rs66&\rs726&\rs1386 \\
        \hline
        Training Loss [$^{\times10^-3}$]&    \textbf{2.70}&2.96&3.04&         &1.25&2.43&2.76&    &\textbf{1.29}&\textbf{2.09}&\textbf{2.81}& &1.18&2.17&3.66\\  
        
                                              
        Validation Loss [$^{\times10^-3}$]&
        \textbf{2.65}&1.65&1.94&         &14.5&2.01&2.64&        &\textbf{27.54}&\textbf{2.76}&\textbf{3.15}&       &45.21&5.02&4.10\\
        \hline
        \end{tabular}
    }
\caption{Loss after training the Network for 30000 iterations. The dataset is composed with 80\%, 50\%, 20\% and 10\% of the initial dataset (527, 330, 132 and 66 samples, respectively) to train the Network using three types of data. (\textbf{O}: Original examples; \textbf{T}: translation and illumination changes and \textbf{O} examples; \textbf{P}: Perlin noise augmentation and \textbf{T} examples, check Section~\ref{sec:augmentation})}
\label{tab:loss} 
\end{table*}

\section{Results}

In this section we present the results of the proposed cleaning architecture showing: i) a detailed analysis of the performance of the neural network and of the TP-GMM correlating it with the amount of training examples used and data augmentation strategies and ii) real-world experiments testing the learned architecture on a different robotic platform - the DoRo robot (see Fig.\ref{fig:doro}).

\subsection{\replaced[]{Validation set}{Test set results}}

The results presented on this section were evaluated on (the same) 20\% of the original iCub cleaning dataset, which we call for now on as \textit{Validation Set}. 
We exploit the 80\% remaining examples to train the network and the TP-GMM algorithm, testing several strategies of: i) data augmentation (for the case of the network) and ii) different quantities of training data to access the amount of necessary examples to achieve a good accuracy in the cleaning task \added[]{(on both, CNN and TP-GMM)}.

\subsubsection{Network tests}
\label{subsec:nettests}
The execution of kinesthetic demonstrations to feed the system with learning examples could be time consuming, so to access the performance of the Network on the validation set according to the data present in the training set, we run the Network several times with different types of data augmentation and with different amounts of initial kinesthetic teaching examples.

\added[]{We have created 12 (different) training sets combining four (4) percentages of the original dataset (80\% - 527 samples, 50\% - 330 samples, 20\% - 132 samples and 10\% - 66 samples) with three (3) data types (\textbf{O}, \textbf{T} and \textbf{P}). The datasets of type \textbf{O} include only the Original samples, the datasets \textbf{T} extends \textbf{O} adding 2 data augmentation strategies (variations of illumination and Translation) and the \textbf{P} datasets include the previous ones adding the Perlin noise images as well. }
The performance of the Network on the validation set taking into consideration the amount of data used and augmentation strategy performed can be seen in Table~\ref{tab:loss}.

\deleted[, remark=obsolete]{The training set was constructed combining: i) the percentage of the initial dataset used (\textit{i.e.}, 80\% , 50\% , 20\%  10\%) and ii) the type of data on the training set (\textit{i.e.}, (\textbf{O}) the original images, (\textbf{T}) adding variations of illumination and translation as data augmentation, and (\textbf{P}) augmenting the dataset with Perlin noise images as well). }

The training loss is similar in all the training sets which implies that the Network is learning (\textit{i.e.}, reducing the error) on those datasets, however, the validation loss increases when we feed the network with less examples. For instance, using only 20\% of the original dataset \added[]{(\textbf{O20\%})}, the loss increases one order of magnitude \added[]{(from 2.65 to 27.54, on \textbf{O80\%} and \textbf{O20\%}, respectively)}. Looking on the data augmentation strategies (\textbf{T} and \textbf{P}) using only 20\% of the available data, one can see that the validation loss is similar to the case the network is trained on 80\% of the original dataset is used \added[]{(\textbf{O80\%} = 2.65; \textbf{T20\%}=2.76 and \textbf{P20\%}=3.15)}. Comparing the most promising networks with data augmentation (\textbf{T20\%} and \textbf{P20\%}) with the networks trained on the original dataset (\textbf{O80\%} and \textbf{O20\%}), one can see the evolution of the loss on training and validation on Fig.~\ref{fig:lossNetworks}. In Fig.~\ref{fig:lossNetworks}, the solid line is the filtered loss signal using a moving average filter with a window size of 5 (\textit{i.e.}, 500 iterations) and the dotted signal is the original (non-filtered) data. \added[]{Furthermore, with only 10\% of the kinesthetic teaching examples (66 samples), the network is not able to generalize well, achieving a validation error 2 times bigger.}

\added[]{Clearly, the Perlin noise is not essential on the validation set (achieving a similar validation loss). This happens because the background is roughly the same (the environment did not change). However, as can be seen in Section 4.2, it will be essential when generalizing to another background (on the DoRo robot).
After this evaluation, we conclude that \textbf{T20\%} and \textbf{P20\%} are suitable to test on the real robot and are a good trade-off between number of kinesthetic teaching and accuracy achieved.}

\subsubsection{TP-GMM tests}

\begin{figure}
\centering
\includegraphics[width=0.46\textwidth]{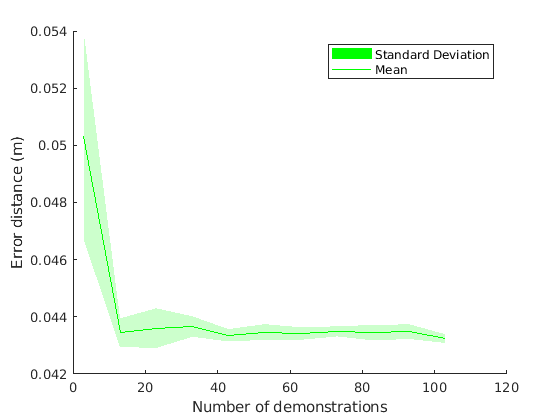}
\caption{Variation of the error between GMR generated trajectories and kinesthetic ones, with the number of demonstrations presented to the GMM. 
}
\label{fig:gmmSel}
\end{figure}

The TP-GMM should be learned using cleaning trajectories as demonstrations. In order to access the amount of demonstrations needed to learn to generate the task trajectories we use the 80\% of the original dataset as training and 20\% of the dataset as validation set \added[]{(the same validation set in Sec.~\ref{subsec:nettests})}. The TP-GMM was initialized using several quantities of random sampled demonstrations and the learned model was tested on the validation set. Fig.~\ref{fig:gmmSel} depicts the mean and standard deviation on 10 trials (increase each 10 demonstrations) of the error between GMR generated trajectories and kinesthetic ones as they vary in the number of demonstrations used.

\begin{table*}
\centering
    \begin{tabular}{l c c c c c}
                    & \multicolumn{2}{c}{\textbf{Marker}} & & \multicolumn{2}{c}{\textbf{Lentils}}\\
    \cline{2-3}
    \cline{5-6}
                    & Area cleaned & Standard Deviation & & Distance reduced & Standard Deviation \\ 
    \hline
    Cauli et al. (iCub) \cite{cauli2018autonomous}& 80\% & 15\% && 45\% & 2\% \\  
    Our results (DoRo) & 75\% & 20\% && 50\% & 10\% \\ 
    \hline
    \end{tabular}
\caption{\added[]{Comparison between the previous results of \cite{cauli2018autonomous} on the iCub robot and our results on the DoRo robot. The test scenario is the same for both the experiments. The two systems have a different network architecture and a different data augmentation strategy.}}
\label{tab:comparison} 
\end{table*}

\subsection{Robot experiments}
\label{sub:robotexp}
\begin{figure}
\centering
 
    \subfloat[r = 1]
    {\includegraphics[width=0.45\linewidth]{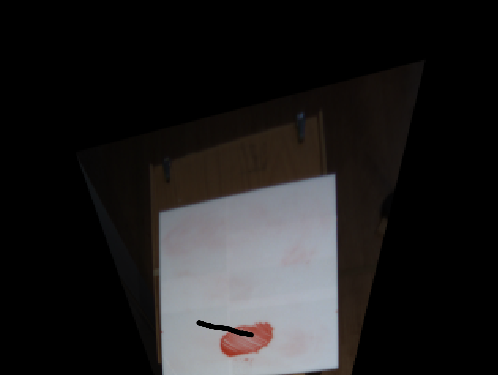}} 
    \subfloat[r = 1]
    {\includegraphics[width=0.45\linewidth]{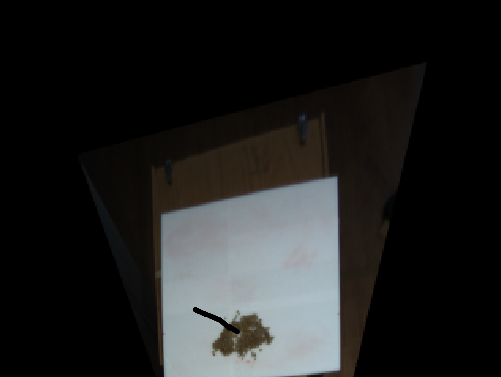}} \\
    \subfloat[r = 2]
    {\includegraphics[width=0.45\linewidth]{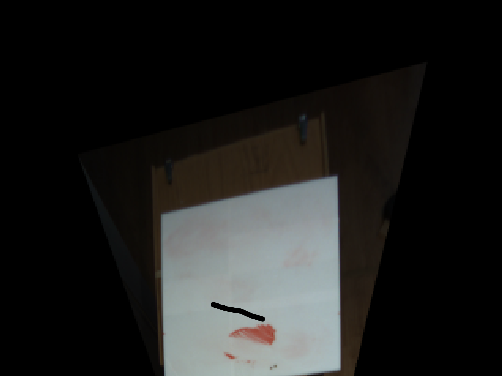}} 
    \subfloat[r = 2]
    {\includegraphics[width=0.45\linewidth]{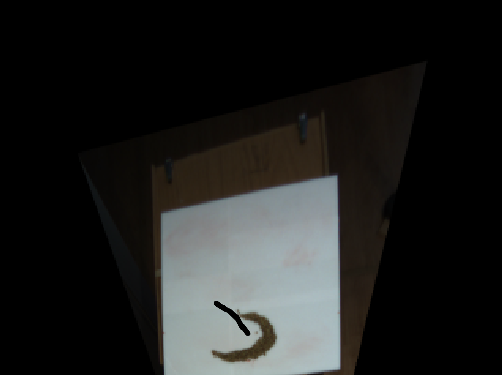}} \\
    \subfloat[r = 3]
    {\includegraphics[width=0.45\linewidth]{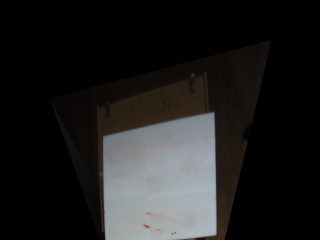}}  
    \subfloat[r = 3]
    {\includegraphics[width=0.45\linewidth]{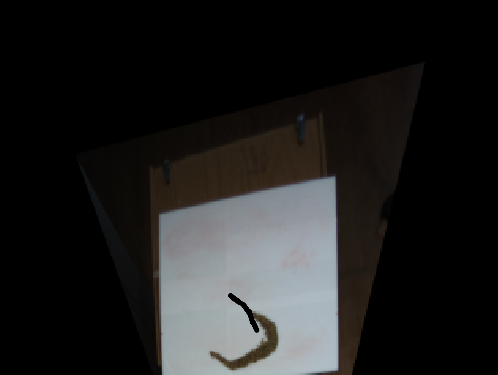}} \\
    \subfloat[r = 4]
    {\includegraphics[width=0.45\linewidth]{images/38_original}} 
    \subfloat[r = 4]
    {\includegraphics[width=0.45\linewidth]{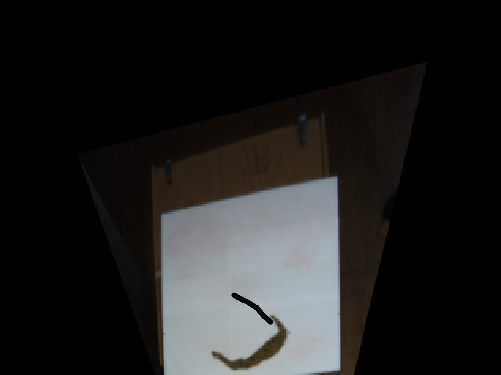}} \\
    \subfloat[r = 5]
    {\includegraphics[width=0.45\linewidth]{images/38_original}}
    \subfloat[r = 5]
    {\includegraphics[width=0.45\linewidth]{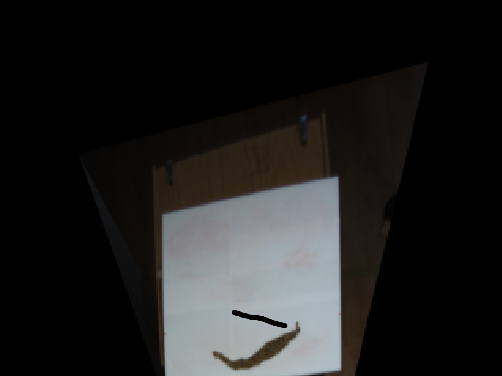}} \\
    
    \subfloat[final result]
    {\includegraphics[width=0.45\linewidth]{images/38_original}} 
    \subfloat[final result]
    {\includegraphics[width=0.45\linewidth]{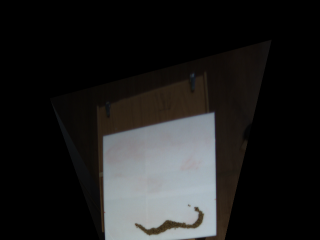}} \\
    \caption{Testing examples on the real robot - DoRo - \added[]{over 5 budget repetitions}. In black color it is possible to see the output trajectory. Left Column: marker scribbles; Right Column: lentils.}
    \label{fig:examples}
\end{figure}

\begin{figure}[t]
\centering
\includegraphics[width=0.46\textwidth]{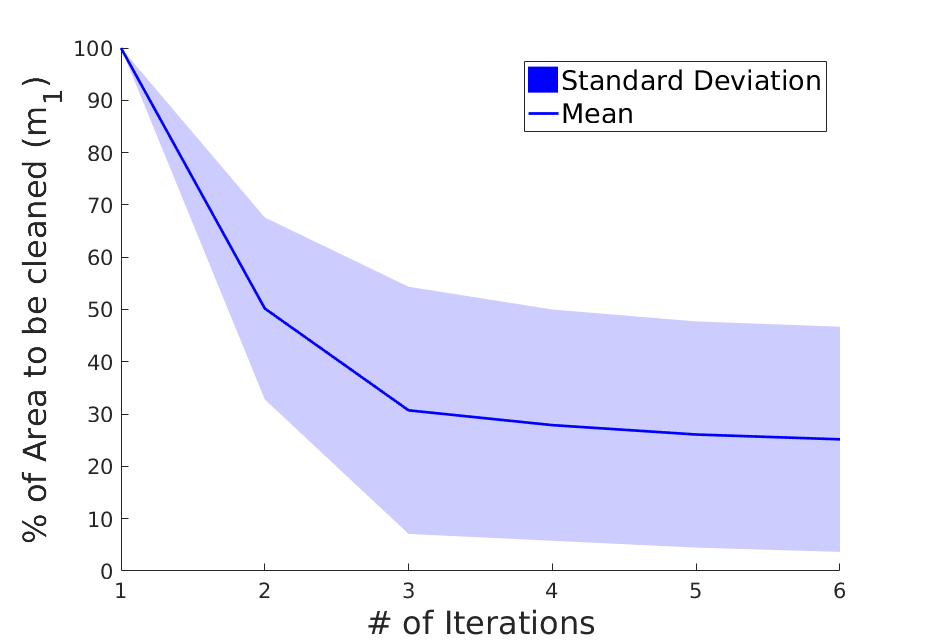}
\caption{Mean and standard deviation, using evaluation metric 1 (See Eq.~\eqref{equ:metric1}), for 15 Marker Experiments with 3 different table height on the DoRo Robot. }
\label{fig:markermean}
\end{figure}

\begin{figure}[t]
\centering
\includegraphics[width=0.46\textwidth]{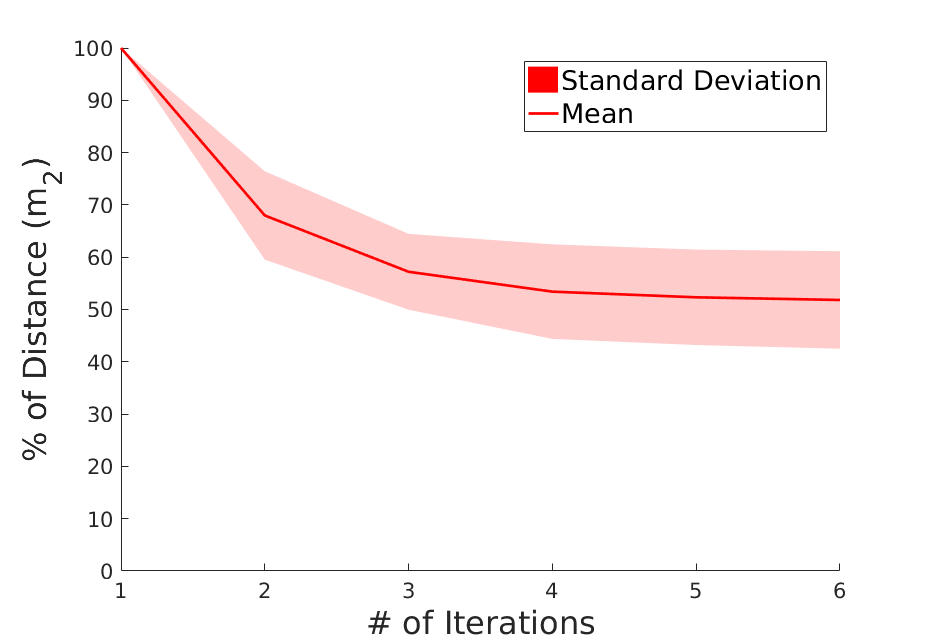}
\caption{Mean and standard deviation, using evaluation metric 2 (See Eq.~\eqref{equ:metric2}), for 15 Lentils Experiments with 3 different table height on the DoRo Robot.}
\label{fig:lentilsmean}
\end{figure}

The proposed architecture was tested on a real scenario using the DoRo robot (See Fig.~\ref{fig:doro}) to determine the transferring capabilities of the cleaning system to a different robotic platform. The robot should try to clean the dirty table (with marker scribbles or cluster of lentils) using a budget of five (5) repetitions ($N_r=5$). For each repetition ($r$), the agent looks to the table, detects the dirt and adjusts the output trajectory accordingly \added[]{(same experimental setup as in \cite{cauli2018autonomous} for the iCub robot)}. In Fig.~\ref{fig:examples}, one example of cleaning markers scribbles (left column) and one of cluster of lentils (right column) can be seen with the output trajectory super-imposed on the image with black color using the P20\% Network. In this case, the red ink was cleaned after the second repetition and the cluster of lentils is closer to the right bottom corner of the table after the five repetitions (final result).
\added[]{Note that, we did not draw the trajectories generated on the marker scribbles inside the five repetitions budget (\textit{i.e.}, $r=3, 4, 5$), since the table was already clean (apart from some small fragments invisible for the robot). Moreover, our architecture does not have a criteria to stop the cleaning task (see Section~\ref{sec:conclusions} for further discussion), so the robot will perform always the same trajectory if an input image with a clean table is shown.}


In a more quantitative analysis, and using the error metric 1 defined on Eq.~\eqref{equ:metric1}, the DoRo robot performed 15 cleaning experiments on marker scribbles setting the table at 3 different heights. The results over the 5 repetitions budget can be seen in Fig.~\ref{fig:markermean}. We reduced the dirt in 75\% of its initial area with a standard deviation of 20\%.
In the lentils case, the table was set at the same 3 different heights and the robot performed 15~different experiments. The mean error and standard deviation using the metric $m_2$ (see Eq~\eqref{equ:metric2}) can be seen in Fig.~\ref{fig:lentilsmean}. The percentage of the initial distance from the bottom right corner of the table (the target point when cleaning this type of dirt) was reduced in 50\% with a standard deviation of 10\%. 

\added[]{Table \ref{tab:comparison} shows the comparison between the results obtained on the DoRo and the results of \cite{cauli2018autonomous} obtained on the iCub (networks and datasets of the 2 systems have some differences. Please refer to sections \ref{sub:contrib} and \ref{sec:Dataset}). The results on the DoRo are close to the results obtained on the iCub, showing how a system trained on one robot can be used to control a second one.}

\replaced[]{The networks trained on \textbf{O80\%}, \textbf{T20\%} and \textbf{P20\%} (check Fig.~\ref{fig:lossNetworks} and Table~\ref{tab:loss})}{The network presented in Fig.~\ref{fig:lossNetworks}, namely networks O80\%, O20\%, T20\% and P20\%,} were tested on the DoRo robot. The network trained with only the original images (\textbf{O80\%} \deleted[]{and O20\%}) was not able to detect and clean any type of dirt. The \textbf{T20\%} network was able to clean the dirt when its location was on the central part of the table but with lower overall performance. Indeed, the data augmentation with Perlin noise (\textbf{P20\%}) was essential for transferring the learned cleaning movements from the iCub to the DoRo robot\added[]{, since the background surrounding the robot is completely different on the DoRo robot (see examples in Fig.~\ref{fig:examples}) and on the iCub (see original dataset on Fig.~\ref{fig:aug} - left).}

\section{Conclusions and Future work}
\label{sec:conclusions}
\replaced[]{We presented a framework for learning how to perform a given cleaning task from human kinesthetic demonstrations, directly from raw camera images, and later transferring the knowledge gathered in this process to a different robot. The parameters of the convolutional neural network trained using the provided demonstrations can be directly used in a different robot if some care is taken to make the network invariant to illumination and perspective changes when applied to a different robot. To achieve this we employ several techniques, such as using a virtual camera to achieve perspective invariance across robots and data augmentation by random changes in illumination, image translation and adding Perlin noise to the background regions of the images. The use of these strategies reduced the need for a large training set --- only 20\% of the recorded data was needed to achieve a similar test error when compared to the situation where no data augmentation was used. Furthermore, the robustness provided by the use of these techniques allowed for a straightforward use of the trained CNN in a different robot: the trained CNN using data acquired from human demonstrations on the iCub robot was used in the DoRo platform to perform the same task without noticeable loss of performance.}{We presented a fully autonomous robotic platform performing cleaning tasks based on a deep neural network architecture using human examples as learning demonstrations. The system proved to be able to generalize well to different environments. Indeed, the demonstrations were performed on the iCub robot and the knowledge transferred to the DoRo platform without losing performance. The data augmentation strategy using translations, illumination changes and Perlin noise with only 20\% of the recorded data proved to be sufficient to increase the robustness of the algorithm.}

\replaced[]{A current limitation of the proposed framework is the need to perform an initial manual calibration to set up the virtual camera on the second robot, so that the learned CNN can be used to perform the cleaning task. Also, currently the trajectory generation is performed in open-loop, with intermediate trajectory points provided by the network learned from human demonstrations. As a consequence, the robot can become stuck performing the same trajectory over and over again when the performed movement does not significantly change the dirt configuration: this typically happens when the table is almost cleaned, as discussed in the previous section}{In future work, we would like to add different types of dirt to make the cleaning task more complete which could be useful for real scenarios. Moreover, we are also planning to develop cleaning tasks using deep reinforcement learning \cite{devin2017deep}, where the agent could use the information about the dirt on its vision sensors as a reward.}
\replaced[]{Implementing a stopping criteria based on the detection of a clean table could be a straightforward solution for this problem.}{One solution could be to implement a stopping criteria detecting a clean table.}
\added[]{Moreover, a direction for further research that can alleviate this issue is to use a deep reinforcement learning approach \cite{devin2017deep}, where the robot can learn from trial and error how to clean the table, based on a set of image features provided by a deep neural network.}

\begin{acknowledgements}
This work was partially supported by Funda\c{c}\~{a}o para a Ci\^{e}ncia e a Tecnologia (project UID\-/\-EEA\-/\-50009\-/\-2013 and Grant PD\-/BD\-/\-135115\-/\-2017) and the RBCog-Lab research infrastructure. We acknowledge the support of NVIDIA Corporation with the donation of the GPU used for this research.

\end{acknowledgements}


\bibliographystyle{spmpsci}
\bibliography{bib/ieeeexample}{}

\end{document}